\documentclass{neus2025}

\def\eqref#1{equation~\ref{#1}}

\def\1{\bm{1}}

\DeclareMathAlphabet{\mathsfit}{\encodingdefault}{\sfdefault}{m}{sl}
\SetMathAlphabet{\mathsfit}{bold}{\encodingdefault}{\sfdefault}{bx}{n}

\usepackage{hyperref}
\usepackage{listings}

\usepackage{enumitem}
\usepackage{import}
\usepackage{mdframed}
\usepackage{nicefrac}
\usepackage{todonotes}
\usepackage{xcolor}
\usepackage{pythonhighlight}
\usepackage{placeins}
\usepackage{transparent}
\newcommand{\heart}{\ensuremath\heartsuit}
\definecolor{blueTileColor}{HTML}{afafff}
\newcommand{\blueTile}{{\color{blueTileColor}\blacksquare}}
\newcommand{\yellowTile}{{\color{yellow}\blacksquare}}
\definecolor{brownTileColor}{HTML}{f4a460} 
\newcommand{\brownTile}{{\color{brownTileColor}\blacksquare}}
\definecolor{redTileColor}{HTML}{ff8b8b} 
\newcommand{\redTile}{{\color{redTileColor}\blacksquare}}
\newcommand{\trc}{\xi}
\newcommand{\alphabet}{\Sigma}

\newcommand{\concat}{\centerdot}
\newcommand{\size}{\text{size}}
\newcommand{\Nat}{\mathbb{N}}
\usepackage{hyperref}
\usepackage{algorithm}

\usepackage[noend]{algorithmic}

\title[L$^*$LM]{L$^*$LM: Learning Automata from Demonstrations, \\Examples, and Natural Language}
\usepackage{times}

\author{%
 \Name{Marcell Vazquez-Chanlatte} \Email{marcell.chanlatte@nissan-usa.com}\\
 \addr Nissan Advanced Technology Center Silicon Valley
 \AND
 \Name{Karim Elmaaroufi} \Email{elmaaroufi@berkeley.edu}\\
 \addr University of California, Berkeley
 \AND
 \Name{Stefan Witwicki} \Email{stefan.witwicki@nissan-usa.com}\\
 \addr Nissan Advanced Technology Center Silicon Valley
 \AND
 \Name{Matei Zaharia} \Email{matei@berkeley.edu}\\
 \addr University of California, Berkeley
 \AND
 \Name{Sanjit A. Seshia} \Email{sseshia@eecs.berkeley.edu}\\
 \addr University of California, Berkeley
}

\begin{document}

\maketitle

\vspace{-1em}
\begin{abstract}
Expert demonstrations have proven to be an easy way to indirectly specify
complex tasks. Recent algorithms even support extracting unambiguous formal
specifications, e.g. deterministic finite automata (DFA), from demonstrations.
Unfortunately, these techniques are typically not sample-efficient. In this
work, we introduce $L^*LM$, an algorithm for learning DFAs from both
demonstrations \emph{and} natural language. Due to the expressivity of natural
language, we observe a significant improvement in the data efficiency of
learning DFAs from expert demonstrations. Technically, $L^*LM$ leverages large
language models to answer membership queries about the underlying task. This is
then combined with recent techniques for transforming learning from
demonstrations into a sequence of labeled example learning problems. In our
experiments, we observe the two modalities complement each other, yielding a
powerful few-shot learner.
\end{abstract}


\section{Introduction}
Large Language Models (LLMs) have emerged as a powerful tool for converting
natural language expressions into structured tasks~\cite[]{yang2023automatonbased,song2023llmplanner,DBLP:conf/icml/HuangAPM22}. 
Similarly, in many settings (e.g. robotics), demonstrations and labeled examples provide a
complementary way to provide information about a task~\cite[]{DBLP:journals/arcras/RavichandarPCB20}.
In addition, natural language has been shown to 
significantly reducing the number of demonstrations needed to learn to perform
a task~\cite[]{roboclip}.
%
Although impressive, these methods suffer a core limitation: they do not
provide a well-defined artifact that \emph{unambiguously} encodes the
specification of the task in a manner that supports: \textbf{(i)} formal
analysis and verification, and \textbf{(ii)} \textit{composition} of tasks.

For example, we may wish to compositionally learn two task specifications
independently in environments that facilitate learning them and compose
them afterwards: ``dry off before recharging'' or ``enter water before
traversing through hot regions''. Similarly, due to regulatory requirements, we
may wish to enforce an additional set of rules conjunctively with the learned
specification, e.g., ``never allow the vehicle to speed when children are
present.'' In both cases, a desirable property of our learned task
representation is that it can guarantee high-level system properties without
retraining. Any need to fine-tune learned tasks with such properties undercuts
the original purpose of learning generalizable task representations
~\cite[]{DBLP:journals/corr/LittmanTFIWM17,DBLP:conf/nips/Vazquez-Chanlatte18}.

To this end, we consider learning task specifications in the form of
deterministic finite automata (DFA). The choice of DFAs as the concept class is
motivated by three observations.
First, DFAs offer simple and intuitive semantics that require only a cursory
familiarity for formal languages. The only requirement to interpret them is a
basic understanding of how to read flowcharts. As such, DFAs offer a balance
between the accessibility of natural language and rigidity of formal semantics.
Second, DFAs explicitly encode memory, making the identification of relevant
memory needed to encode the task clear. Furthermore, they are the ``simplest''
family of formal languages to do so, since they are equivalent to having finite
number of residual languages (Nerode congruences in the form of
states)~\cite[]{DBLP:books/aw/HopcroftU79}.
Third, many existing formulations such as finite temporal logic and sequences
of reach avoid tasks (go to location A, while avoiding B, then go to C
while avoiding D) are regular languages and thus are expressible as
DFAs~\cite[]{DBLP:conf/kr/CamachoBM18}.

\begin{figure*}[htbp]
    \centering
    \scalebox{1.25}{
      \includegraphics{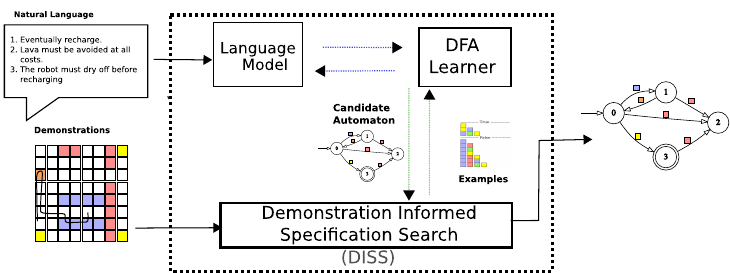}
    }
    \vspace{-1em}
    \caption{$L^*LM$ is a multi-modal learning algorithm that builds upon the classic $L^*$ automata learning algorithm to learn automata from natural language. We also incorporate DISS to learn from expert demonstrations and self-labeled examples that minimize surprisals.\label{fig:overview}}
    \vspace{-0.75em}
\end{figure*}

In this work, we offer three key insights that enable us to robustly learn DFAs
from a mixture of demonstrations, labeled examples, and natural language. The
resulting algorithm is known as $L^*LM$. The first major insight is that we
design an interaction protocol built around answering simple membership
queries. Here, we build on the classic automata learning literature to create
an active learning algorithm that only asks the LLM membership queries, e.g.,
``is it ok to visit the red tile and then the blue tile?'' There are two
natural ways to realize this: (i) asking the LLM to synthesize code and then
evaluating the code and (ii) by constraining the output of the LLM to conform
to a grammar~\cite[]{llamacpp} -- allowing trivial interpretation of the
membership label. In this work, we focus on the latter to avoid arbitrary code
execution but demonstrate in appendix~\ref{sec:llm_code_oracle} that models
such as GPT-4o are capable of generating membership answering programs.
Importantly, as $L^*LM$ is designed to take in user inputs, avoiding arbitrary
code execution removes whole classes of security vulnerabilities due to code
injection.

Our second key insight is that it is important for the LLM to be able to say
it is unsure when asked a membership query. We found in our
experiments that the LLM would often state that it was unsure during chain-of-thought
reasoning~\cite[]{DBLP:conf/nips/Wei0SBIXCLZ22} followed by a hallucinated
membership response. We note that this is consistent with other results in the
literature~\cite[]{DBLP:journals/corr/abs-2305-04388}. The resulting DFA would
then contain features that were not justified by the labeled examples or the
language prompt and resulted in poor alignment with the task. 
A simple solution
was to allow the LLM (or code generated by an LLM as in the appendix) to respond ``unsure'' 
and then have the DFA learner
ask a different query. As shown in our experiments, this greatly
improved performance, particularly when complemented by inferences
made by analyzing the demonstrations. 
Note, the ``unsure escape hatch'' is required because
the LLM is given an incomplete context to specify the task.
For example, parts of the task may be omitted by the user.
In our experiments
\emph{environment dynamics}
are withheld from the LLM. 

This leads to our third key insight: Labeled examples offer a bridge between
LLM knowledge distillation and an outside verifier.
In our experiments, we found that LLMs
such as GPT3.5-Turbo and GPT4-Turbo failed
to correctly provide membership queries
for simple languages. In order to correct
this, we leverage (as a blackbox) the recent Demonstration Informed Specification Search
(DISS) algorithm which translates the problem of learning a DFA from an expert
demonstration in a Markov Decision Process into an iterative series of DFA
identification from labeled example
problems~\cite[]{Vazquez-Chanlatte:EECS-2022-107}. Each problem is sent
to our LLM based DFA-learner -- seeded with the labeled examples as context.
Because of the ability to say its unsure, the LLM is able to focus on labeling
queries it is confident in due to the text prompt and leverages DISS to provide corrective feedback.
By creating an interaction protocol with DISS, we fuse the LLM's natural
language reasoning with dynamics-dependent
analysis the LLM would otherwise be oblivious to.

\paragraph{Contributions:}
\vspace{-0.6em}
\begin{enumerate}[noitemsep,leftmargin=1.3em]
    \vspace{-1em}
    \item We propose $L^*LM$, a novel algorithm for multimodal
        learning of deterministic finite automata from (i) natural language,
        (ii) labeled examples, and (iii) expert demonstrations in a Markov
        Decision Process.

    \item A prototype implementation of $L^*LM$ written in
        Python and compatible with many LLMs.
        \footnote{Our code will be released after review.}

    \item We empirically illustrate that (i) providing a natural language description      of the task improves learnability of the underlying DFA, (ii) allowing the language model to respond unsure improves performance, and (iii)
    allowing more queries to the language
        model improves performance.

\end{enumerate}

\noindent
\vspace{-0.25em}
We emphasize that the resulting class of algorithms:
\begin{enumerate}[noitemsep,leftmargin=1.5em]
  \vspace{-0.8em}
  \item is guaranteed to output a valid DFA that is consistent with the input examples.
  \item requires no arbitrary code evaluation.
  \item only asks simple yes/no/unsure questions to the LLM.
  \item supports natural language task descriptions and a-priori known examples and demonstrations.
\end{enumerate}

\section{Related Work}

\vspace{-0.5em}
This work lies at the intersection of a number of fields including grammatical
inference, knowledge distillation from language models, and multi-modal
learning. We address these connections in turn.

\textbf{Grammatical inference and concept learning: }
Grammatical inference~\cite[]{de2010grammatical} refers to the rich literature on
learning a formal grammar (often an automaton ~\cite[]{DBLP:conf/tacas/DrewsD17,DBLP:journals/corr/abs-1008-1663}) from data -- typically labeled
examples. Specific problems include finding the smallest automata consistent
with a set of positive and negative strings~\cite[]{de2010grammatical} or
learning an automaton using membership and equivalence
queries~\cite[]{DBLP:journals/iandc/Angluin87}. 
We refer the reader to~\cite[]{DBLP:conf/fmcad/Vaandrager21} for a detailed
overview of active automata learning.
Notably, we leverage SAT-based
DFA-identification~\cite[]{DBLP:conf/lata/UlyantsevZS15,DBLP:conf/icgi/HeuleV10}
to easily identify small DFAs that are consistent with a set of labeled
examples and utilize a common technique to convert this passive learner into an active
version space learner~\cite[]{DBLP:conf/ictai/SverdlikR92}. Our LLM to DFA extraction
pipeline builds directly on these techniques.
Finally, we note that the idea of learning with incomplete teachers is an evolving topic in automata learning~\cite[]{DBLP:conf/ecoop/MoellerWSKF023}.

\textbf{Learning from expert demonstrations. }
The problem of learning objectives by observing an expert also has a rich and well
developed literature dating back to early work on Inverse Optimal
Control~\cite[]{kalman1964linear} and more recently via Inverse Reinforcement
Learning (IRL)~\cite[]{DBLP:conf/icml/NgR00}. 
While powerful, traditional IRL provides no principled mechanism for composing
the resulting reward artifacts and requires the relevant historical features
(memory) to be apriori known. Furthermore, it has been observed that small
changes in the workspace, such as moving a goal location or perturbing transition
probabilities, can change the task encoded by a fixed
reward~\cite[]{DBLP:conf/nips/Vazquez-Chanlatte18,DBLP:journals/nips/Abel22}.

To address these deficits, recent works have proposed learning Boolean task
specifications, e.g. logic or automata, which admit well defined compositions,
explicitly encode temporal constraints, and have workspace independent
semantics~\cite[]{DBLP:conf/cdc/KasenbergS17,DBLP:conf/rss/ChouOB20,DBLP:conf/nips/ShahKSL18,DBLP:conf/icra/Yoon021,DBLP:conf/cav/Vazquez-Chanlatte20}. 

Our work utilizes the Demonstration Informed Specification Search 
(DISS) algorithm~\cite[]{Vazquez-Chanlatte:EECS-2022-107}.
DISS is a variant of maximum causal entropy IRL that 
recasts learning a specification (here a DFA) as a series of
grammatical inference from labeled example queries. This is done by analyzing
the demonstration and computing a proxy gradient over the surprisal
(negative log likelihood) which suggests paths that should have
their labeled changed to make the demonstrations more likely. The key insight in our work is that this offers a bridge to our LLM extraction 
formalism by having the LLM fill in
some examples and having DISS provide the others. From the perspective
of DISS, this can be seen as indirectly restricting the concept class using
a natural language prompt.

\textbf{Knowledge Extraction from LLMs. }
Attempts to extract formal specifications from language can at least be traced
back to~\cite[]{DBLP:journals/cj/VaderaM94}, if not further. More recent works
have studied extracting knowledge from deep learning models, e.g., extracting an
automaton from a recurrent neural network using $L^*$~\cite[]{pmlr-v80-weiss18a}
or using language prompts to synthesize
programs~\cite[]{DBLP:conf/icse/DesaiGHJKMRR16}. 
Several recent works have explored extracting finite state automata from large
language models~\cite[]{yang2023automatonbased,yang2023multimodal}.

A key difference between our work and the ones mentioned is our focus on multimodal learning from
demonstrations \emph{and} language. Further, as mentioned in the introduction,
when compared against program synthesis, a key feature of the work is the
restriction of arbitrary code to membership queries which avoids (i) security and analysis issues
and (ii) guarantees the resulting concept is always a valid DFA, faithful to the input examples. 
This latter point is a significant difference from other automata extraction works,
e.g.,~\cite[]{yang2023automatonbased}, that focus on more direct extraction of the
automata as a series of steps. Further, we note that the description of steps
(i) encodes a policy rather than a task specification as studied in this paper
(ii) is ultimately restricted to automata whereas the key ideas of our
technique are applicable to arbitrary formal language learners.

Finally, we note that the fact that LLMs hallucinate on unknown queries is well
known~\cite[]{DBLP:journals/corr/abs-2305-04388}, with some works going so far as
to retrain the LLM to refuse answering such
queries~\cite[]{DBLP:journals/corr/abs-2311-09677}. In our work, we do not
retrain but instead relax our membership queries to allow the LLM to say
``unsure'', using constrained
decoding~\cite[]{DBLP:conf/naacl/TrombleE06,DBLP:conf/emnlp/GengJP023}.

\begin{figure}[t]
    \centering
    \subfigure{
        \label{fig:gw8x8}
        \scalebox{0.75}{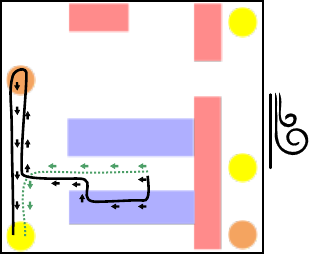}
    }%
    \hspace{1em}
    \subfigure{
        \label{fig:target-dfa}
        \centering
        \scalebox{0.4}{
\begingroup%
  \makeatletter%
  \providecommand\color[2][]{%
    \errmessage{(Inkscape) Color is used for the text in Inkscape, but the package 'color.sty' is not loaded}%
    \renewcommand\color[2][]{}%
  }%
  \providecommand\transparent[1]{%
    \errmessage{(Inkscape) Transparency is used (non-zero) for the text in Inkscape, but the package 'transparent.sty' is not loaded}%
    \renewcommand\transparent[1]{}%
  }%
  \providecommand\rotatebox[2]{#2}%
  \newcommand*\fsize{\dimexpr\f@size pt\relax}%
  \newcommand*\lineheight[1]{\fontsize{\fsize}{#1\fsize}\selectfont}%
  \ifx\svgwidth\undefined%
    \setlength{\unitlength}{488.47432709bp}%
    \ifx\svgscale\undefined%
      \relax%
    \else%
      \setlength{\unitlength}{\unitlength * \real{\svgscale}}%
    \fi%
  \else%
    \setlength{\unitlength}{\svgwidth}%
  \fi%
  \global\let\svgwidth\undefined%
  \global\let\svgscale\undefined%
  \makeatother%
  \begin{picture}(1,0.47297901)%
    \lineheight{1}%
    \setlength\tabcolsep{0pt}%
    \put(0,0){\includegraphics[width=\unitlength,page=1]{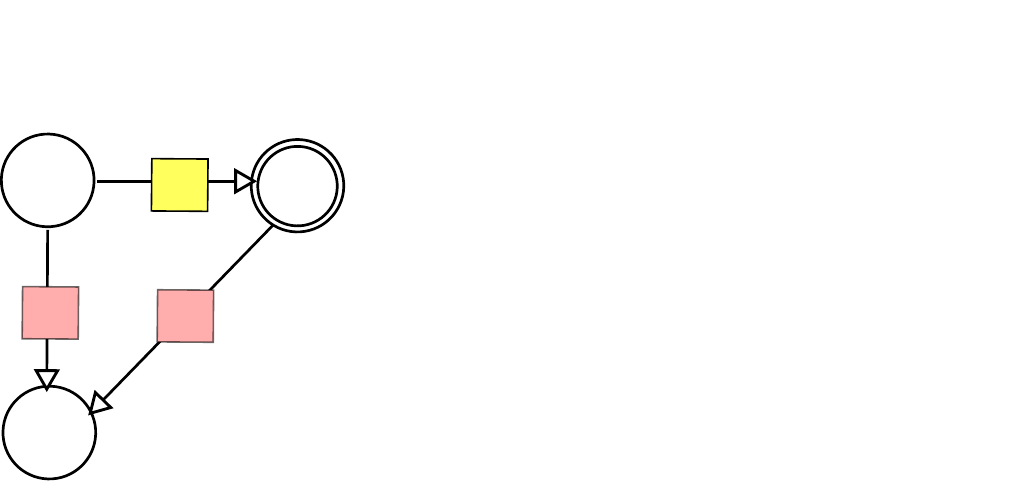}}%
    \put(0.04670961,0.29045387){\makebox(0,0)[t]{\lineheight{1.25}\smash{\begin{tabular}[t]{c}start\end{tabular}}}}%
    \put(0.70917231,0.29016491){\makebox(0,0)[t]{\lineheight{1.25}\smash{\begin{tabular}[t]{c}start\end{tabular}}}}%
    \put(0,0){\includegraphics[width=\unitlength,page=2]{true_dfa.pdf}}%
  \end{picture}%
\endgroup%
}
    }
    \vspace{-0.5em}
    \caption{\ref{fig:gw8x8} Two demonstrations of an example task. While one
        successfully navigates to $\yellowTile$, the other demonstrates an
        implicit rule which we have omitted: dry off before recharing.
        \ref{fig:target-dfa} DFAs with stuttering semantics. If a transition is
        not provided, a self-loop is assumed. Accepting states are marked with
        a concentric circle, and the initial state is labeled start}
    \label{fig:enter-label}
    \vspace{-0.75em}
\end{figure}

\textbf{Running Example. }
To ground our later discussion, we develop a running example.
This running example is adapted from~\cite[]{Vazquez-Chanlatte:EECS-2022-107}.
Consider an agent operating in a 2D workspace as shown in
Fig~\ref{fig:gw8x8}. The agent can attempt to move up, down, left, or
right, but with probability \nicefrac{1}{32}, wind will push the agent
down, regardless of the agent's action. The agent can sense
four types of tiles: red/lava ($\redTile$), blue/water ($\blueTile$), yellow/recharging ($\yellowTile$), and brown/drying ($\brownTile$).

We would like to instruct the robot to \textbf{(i)} avoid lava and \textbf{(ii)} eventually go
to a recharge tile. To communicate this, we provide a variant
of that natural language task description. Further, we provide a few
demonstrations of the task as shown in Fig~\ref{fig:gw8x8}.
Unfortunately, in providing the description, we forget to mention one
additional rule: if the robot gets wet, it needs to dry off before
recharging.

An insight of our work is that the demonstrations provided
imply this rule. In particular, the deviation after slipping implies
that the direct path that doesn't dry off is a negative example.
As we show in our experiments, neither the demonstrations nor the
natural language alone is enough to consistently guess the correct DFA
(shown on the right in Fig~\ref{fig:target-dfa}.

\vspace{-0.25em}
\section{Refresher on Automata Learning}
\vspace{-0.25em}
In this section, we propose a general scheme for extracting a DFA from a large
language model (LLM) that has been prompted with a general task description. We
start with the definition of a DFA and a refresher on automata learning using examples
and membership queries. This then sets us up for an interactive protocol
with the LLM to extract the underlying DFA. Again, we start with the formal
definition of a DFA and a set of labeled examples.

\vspace{-0.5em}
\begin{definition}
A \textbf{Deterministic Finite Automaton} (DFA) is a 5-tuple,
$D = \langle Q, \alphabet, \delta, q_0, F\rangle$, where $Q$ is a finite set of
\textbf{states}, $\alphabet$ is a finite alphabet,
$\delta : Q \times \alphabet \to Q$ is the \textbf{transition
function}, $q_0 \in Q$ is the \textbf{initial state}, and $F \subseteq Q$
are the \textbf{accepting states}. The transition function is lifted
to strings, $\delta^* : Q \times \alphabet^* \to Q$. One says $D$
\textbf{accepts} $x$ if $\delta^*(q_0, x) \in F$.  Denote by $L[D] \subseteq \alphabet^*$
the set of strings, i.e., \textbf{language}, accepted by $D$. Its complement, the set
of \textbf{rejecting} strings, is denoted by $\overline{L[D]}$.
A word, $x$, is said
to \textbf{distinguish} $D_1$ and $D_2$ if it causes one to accept and the other reject, i.e., it 
$x$ lies in the symmetric difference of their languages,
$x \in L[D_1] \ominus L[D_2]$. Finally, we will assume the DFA is endowed with
a \textbf{size} (or complexity), mapping it to a positive real number. This will
typically encode the number of bits or states needed to represent the DFA.
\end{definition}
\vspace{-0.5em}

A collection of labeled examples, $X = (X_+, X_-)$ is a finite mutually exclusive set
of words where $X_+$ and $X_-$ are called the \textbf{positive} examples
and \textbf{negative} examples respectively. A DFA, $D$, is said to be consistent with
$X$ if it accepts all positive examples and rejects all negative examples, i.e.,
$X_+ \subseteq L[D] \wedge X_- \subseteq \overline{L[D]}$
The \textbf{DFA identification problem} asks to find $k\in\Nat$ DFAs of minimal size
that are consistent with a set of labeled examples.

The DFA identification problem is extremely underdetermined in
general due to there being a countably infinite number of consistent DFA for
any finite set of labeled examples~\cite[]{DBLP:journals/iandc/Gold67}. Moreover, for common
size measures such a number of states, the identification problem is known to
be NP-Hard~\cite[]{DBLP:journals/iandc/Gold78}. Nevertheless, many SAT-based implementations exist which,
in practice, are able to efficiently solve the DFA identification
problem~\cite[]{DBLP:conf/icgi/HeuleV10,DBLP:journals/corr/UlyantsevZS16}.

Notably, this all assumes a \textbf{passive} learner, i.e., one where the
example set $X$ is a-priori provided. Alternatively, one can consider
\textbf{active} learners that directly query for labels. Formally, one assumes
that there is some unknown language $L^*$ and an oracle that can answer queries
about $L^*$. The common model, referred to as the Minimally Adequate Teacher
(MAT)~\cite[]{DBLP:journals/iandc/Angluin87} assumes access to two types of queries: (i)
\textbf{membership} queries, $M(x) = x \in L^*$, and (ii) \textbf{equivalence}
queries, $E(D) = (l, x)$ where $l \in \{0, 1\}$ indicates if $L[D] \equiv L^*$. If $l=0$, i.e., the candidate DFA is incorrect, then
$x$ is a distinguishing string.

The classic algorithm for learning under a MAT is a also called
$L^*$~\cite[]{DBLP:journals/iandc/Angluin87}. $L^*$ is known to perform a polynomial number of membership
queries and a linear number of equivalence queries. Unfortunately, in practice
the equivalence queries are often not realizable, and thus are often approximated
by \textbf{random sampling} or \textbf{candidate elimination}. In the former,
one labels random words from a fixed distribution over words yielding a
probably approximately correct (PAC) approximation of the underlying
language~\cite[]{DBLP:journals/iandc/Angluin87}. In the latter, one uses DFA identification to find a set of
consistent DFAs and queries distinguishing sequences. The guarantee in
candidate elimination is then that $\size(D)$ will be minimized. Note that that
this leads to a folk algorithm for transforming any passive DFA identification
algorithm into an active one. An example of the pseudo code for such an algorithm,
\texttt{guess\_dfa\_VL}, is provided in~Alg~\ref{alg:version_space_learner} (located in the appendix) where \texttt{find\_minimal\_dfas}
refers to an arbitrary DFA identification algorithm.\footnote{The VL here
stands for version space learning of which this algorithm can be seen as an instance of.}

Finally, we observe that in many cases -- as will be the case with our LLMs --
the oracle may not be able to confidently provide membership queries. For
example, if the task description provided does not cover the case provided, the
LLM may be simply hallucinating the membership label. To address this, we
propose an \textbf{extended membership} query that answers true, false, or
unsure. Further observe that~Alg~\ref{alg:version_space_learner} applies in the
extended setting by automatically ignoring unsure responses. This is in contrast
to the $L^*$ algorithm which to our knowledge has no extension to support unsure
responses. For our experiments with $L^*$ we map unsure responses to membership queries to false.

\vspace{-0.25em}
\section{Extracting DFAs from LLM Interaction}\label{sec:dfa-extraction}
\vspace{-0.25em}

\begin{figure}
  \centering
    \scalebox{0.85}{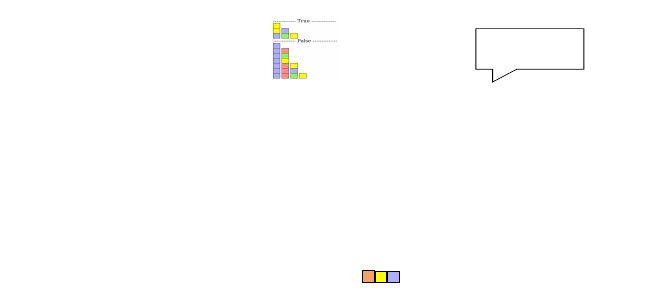}
    \vspace{-0.75em}
    \caption{Visualization of the process of converting a language model
    into a membership oracle for DFA learning\label{fig:llm_oracle}. To address the issue of hallucinations in LLMs, our work proposes an extended membership query and a cache. Ablations of our design are provided in \ref{sec:experiments}}
    \vspace{-0.75em}
  \end{figure}

The previous section formulated automata learning in the passive setting and in
the active setting. Notably, the active setting resulted in interaction protocols
between the learner and an oracle that can answer membership queries. In this
section, we formalize the observation that language models offer a natural
approximation for a membership oracle as overviewed
in~Fig~\ref{fig:llm_oracle}.

Formally, a language model takes in a sequence of tokens, $x \in \Gamma^*$
called a \textbf{prompt}, and outputs another sequence of tokens $y \in
\Gamma^*$ called a \textbf{response}. The response, $y$, can be viewed as a
sample on the suffixes of $x$ from some underlying distribution. Techniques
such as constrained decoding~\cite[]{DBLP:conf/naacl/TrombleE06} offer the ability to further restrict
$y$ to a specified formal language $L_y$ -- for example using a context free
grammar~\cite[]{DBLP:conf/emnlp/GengJP023,llamacpp}. This then guarantees that $y$ can be interpreted as
an extended membership oracle. 
Functionally, our grammar splits the LLM's response into two parts, i.e., 
$y = \text{work}\concat\text{``FINAL\_ANSWER: ''} \concat\text{answer}.$
The work portion can be utilized to implement various prompt engineering tricks,
e.g., ReAct~\cite[]{yao2023reactsynergizingreasoningacting} and Chain-of-Thought~\cite[]{DBLP:conf/nips/Wei0SBIXCLZ22} 
or as seen with reasoning models~\cite[]{openaio1}, space for reasoning tokens. The answer part
contains either yes, no, or unsure -- making parsing into a membership response
trivial.

During the interaction between the LLM and the DFA learner, the prompt is
incrementally extended to include the responses of the previous query. The
initial prompt is taken to be an arbitrary user-provided description of the
task along with (i) instructions for answering the question, (ii) a request to
show work, and (iii) known labeled examples. The membership query is realized by the text: ``is $\langle$ insert word $\rangle$ a positive example?''

Finally, to \emph{guarantee} that the LLM is consistent with the provided
labeled examples, we introduce a caching layer between the LLM and the
DFA learner. This layer answers any known queries without consulting the LLM; 
furthermore, it memorizes any queries the LLM has already answered.

\vspace{-0.25em}
\section{Incorporating Expert Demonstrations}
\vspace{-0.25em}

Next, we discuss how to introduce additional modalities.
\emph{Our key insight is that labeled examples offer a
flexible late stage fusion mechanism between modalities.}
For example, we leverage the Demonstration Informed
Specification Search (DISS) algorithm~\cite[]{Vazquez-Chanlatte:EECS-2022-107} which transforms the problem
of learning concepts from expert demonstrations into
a series of passive learning from labeled example problems.

Specifically, by an expert demonstration we mean the behavior
of an agent acting in a Markov Decision Process who generates
a path, $\xi$, to satisfy an objective. Here, the path is featurized
into a finite alphabet and the goal is to generate a path
that is accepted by some DFA $D$.

\begin{remark}
    We observe several important differences between demonstrations and the
    labeled examples from the previous section. First, a demonstration of $D$
    need not be accepted by $D$. That is, the labeled few-shot examples provided 
    to the LLM must be consistent with the DFA, $D$ whereas the demonstrations 
    need not be. Returning to our running example, this might be
    because (i) the robot slips and accidentally violates the task specification;
    or (ii) the demonstration is a prefix of the final path, perhaps being
    generated in real time.
\end{remark}

The key idea of DISS is to generate counter-factual labeled
examples in a manner that makes the demonstrations less
surprising.
Next, we discuss how to infer a DFA given a collection of expert demonstrations. 
Returning to our running example, if the current candidate DFA encodes \textit{eventually reaching yellow},
then in step (ii) DISS will hypothesize that $\redTile\yellowTile$ is a negative example since it makes
it \emph{less surprising} that the agent didn't take the shortcut through the lava to recharge.
The idea of surprisal is formalized as 
the description complexity (or \textbf{energy}) of the DFA plus the demonstration given its corresponding
policy, 
\[
  U(D, \trc) = -\log \Pr(\trc \mid D) + \lambda\cdot \size(D),
\]
where $\lambda$ is set based on the a-priori expectation of $\size(D)$.
DISS's goal is to use the counter-factual labeled examples to learn new DFAs
that lower $U(\bullet, \trc)$.



Specifically, for each iteration of DISS we run our DFA extraction algorithm from Sec~\ref{sec:dfa-extraction}
with a fixed query budget and split between random and candidate elimination queries. The resulting
DFA is then ranked according to $U$. The processes is repeated for a fixed number of steps and
the DFA with minimal energy $U$ is returned.
Pseudo-code for DISS integration is provided in appendix~\ref{ref:appen_psuedo_code}.
\vspace{-0.5em}

\begin{mdframed}[nobreak=true]
    Our algorithm, $L^*LM\heart\text{DISS}$ uses DISS as a supervisory signal and as a method to incorporate expert demonstrations as an additional learning modality. This enables:

    \begin{enumerate}[noitemsep,leftmargin=1.5em]
    \vspace{-1em}
        \item Resolving ambiguities in the natural language using demonstrations.
        \item Indirectly restricting the concept class of DISS using a natural language prompt.
    \end{enumerate}
\end{mdframed}
\vspace{-0.3em}

Our interaction protocol with DISS is as follows.
For each iteration of DISS, we run our DFA extraction algorithm from Sec~\ref{sec:dfa-extraction}
with a fixed query budget and split between random and candidate elimination queries. The resulting
DFA is then ranked according to $U$. The processes is repeated for a fixed number of steps and
the DFA with minimal energy $U$ is returned.
Pseudo-code for DISS integration is provided in appendix~\ref{ref:appen_psuedo_code}.
\vspace{-0.75em}

\section{Experiments}
\label{sec:experiments}

\begin{figure}[t]
    \centering
    \subfigure{
       \label{fig:lstar_results}
       \centering
       \includegraphics[width=0.48\linewidth]{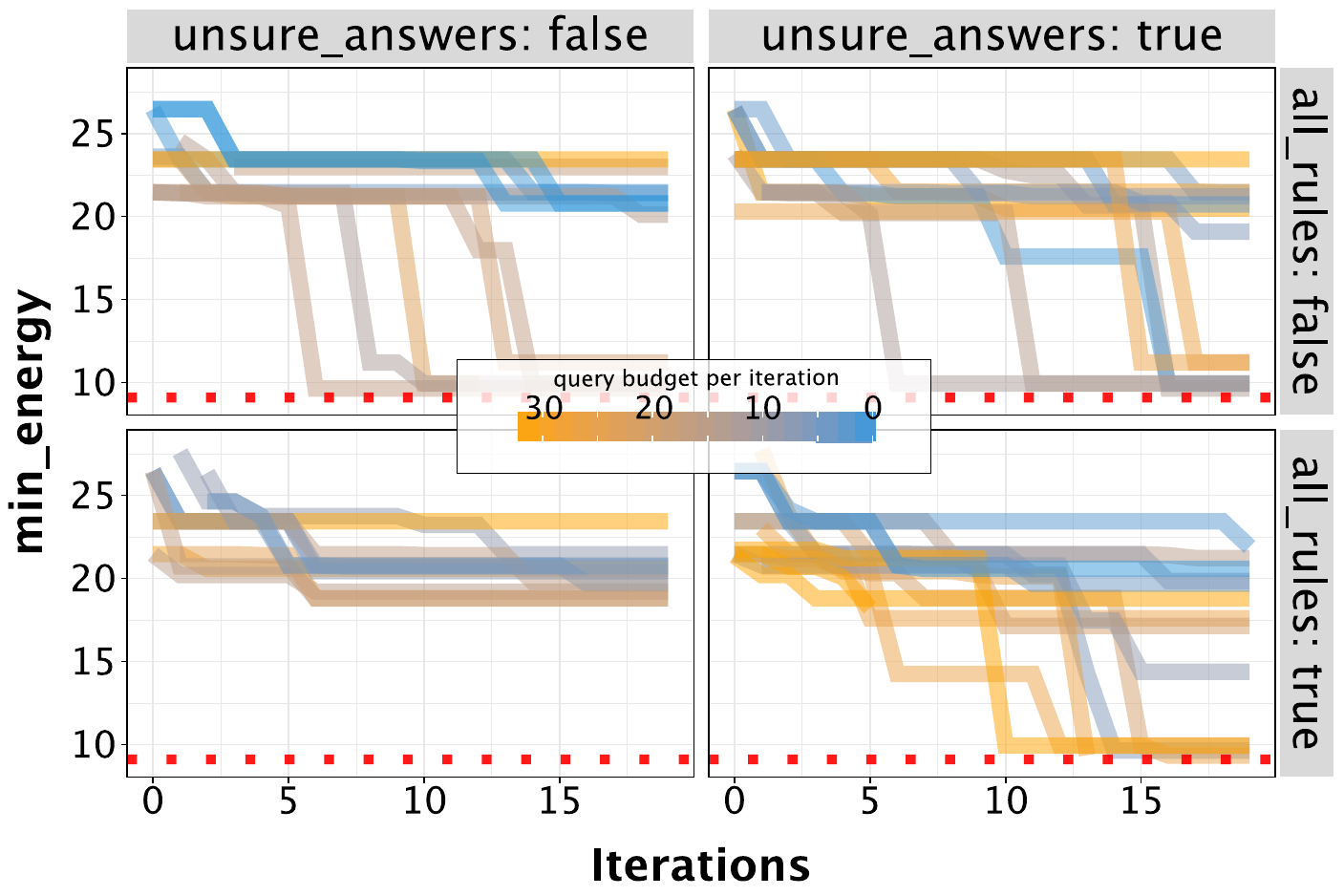}
   }
   \subfigure{
     \label{fig:dfa_identify_results}
     \centering
     \includegraphics[width=0.48\linewidth]{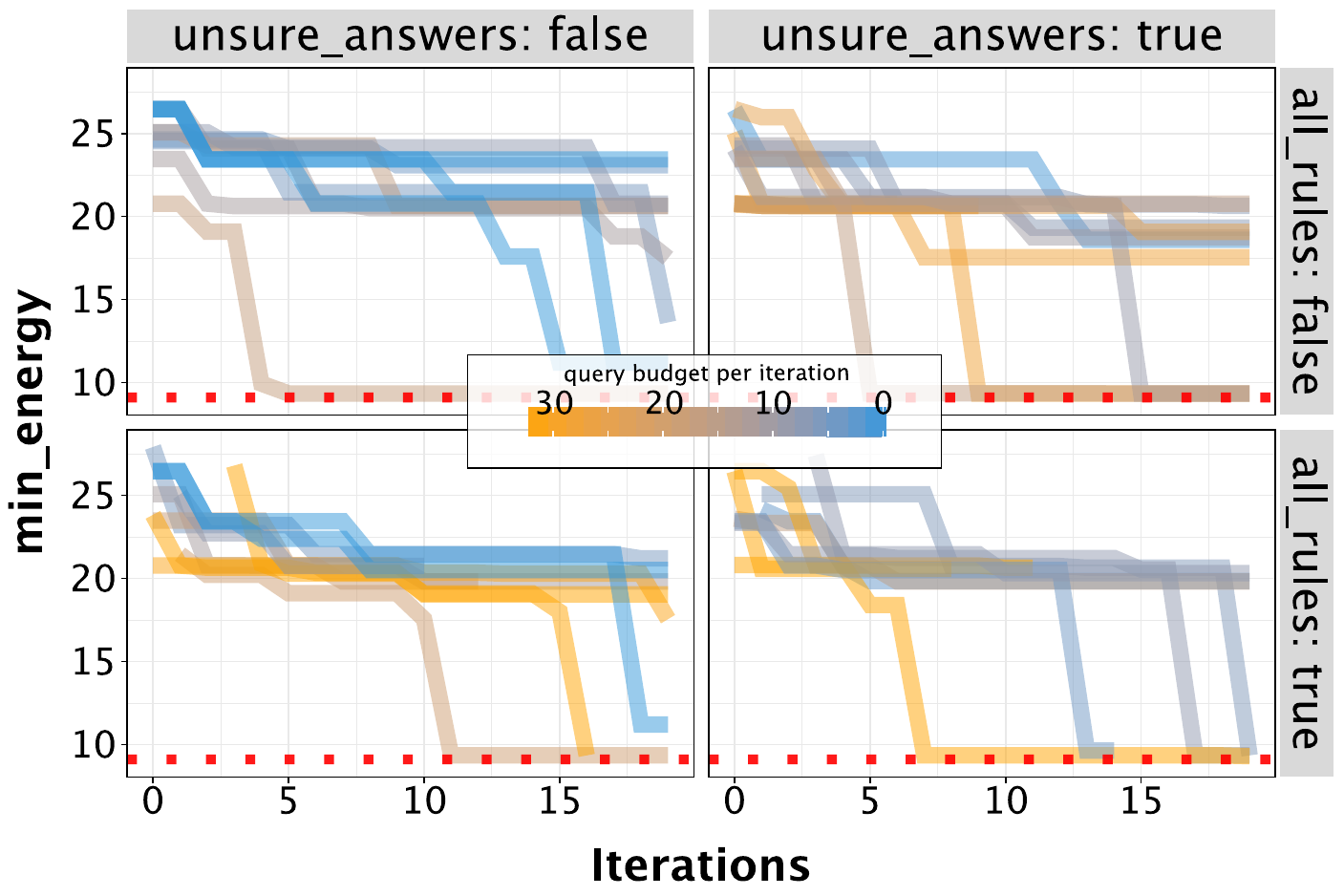}
   }
   \caption{Comparison between \textit{L$^*$ backend} and \textit{candidate elimination variant} in learning DFAs.
        \ref{fig:lstar_results} Running example using \textit{L$^*$ backend}. Ground truth energy is denoted by the red dotted line. Unlike the SAT-based backend (Fig~\ref{fig:dfa_identify_results}), L$^*$ struggles to learn the DFA when not provided all examples and unsure responses are not allowed.
        \ref{fig:dfa_identify_results} Running example using \textit{candidate elimination variant}. In each iteration, DISS conjectures new labeled examples based on the demonstration and the conjectured DFA.
        }
    \label{fig:comparison_results}
    \vspace{-0.75em}
\end{figure}

\label{sec:experiments}

For our first experiment, we use $L^*LM$ as a membership oracle to learn DFAs representing the 7 Tomita Grammars. These simple languages are a common benchmark for studying automata learning~\cite[]{pmlr-v80-weiss18a}. 
We refer the reader to appendix \ref{ref:tomita_exps} for full details of the experiment.
Our results indicate that while \texttt{unsure} does help to reduce eliminations, it cannot eliminate them, so we now switch our focus to the running example where we consider learning through an additional modality (demonstrations) as a way to satisfy the need for a grounded supervisory signal.

\subsection{Robotic Workspace}

Given our findings in the single mode learning setting, we study the following research questions:
\begin{itemize}[noitemsep,labelindent=1.3em,leftmargin=1em]
    \vspace{-0.75em}
    \item \textbf{RQ1:} Does including a \emph{natural language} description of the task aid in inferring the task?
    \item \textbf{RQ2:} Is reasoning using the demonstrations given the natural language prompt important?
    \item \textbf{RQ3:} Does including a \emph{partial} description of the task aid in inferring the task? 
    \item \textbf{RQ4:} Does the LLM avoid hallucinating unhelpful task components when it can say unsure?
    \item \textbf{RQ5:} Does $L^*$ or \textit{dfa-identify} (our SAT-based version space DFA learning) perform better?
\end{itemize}

We adapt the experiment
from~\cite[]{Vazquez-Chanlatte:EECS-2022-107}~Ch~5 which also serves as our running example to support natural language
task descriptions. 
We use the same DFA-conditioned maximum entropy planner as~\cite[]{Vazquez-Chanlatte:EECS-2022-107} which works on a discretized approximation of the 2D workspace.
For a language model, we use Mixtral-8x7B-Instruct~\cite[]{DBLP:journals/corr/abs-2401-04088} and \texttt{find\_minimal\_dfa} is implemented using the \textit{dfa-identify} SAT based solver~\cite[]{Vazquez-Chanlatte_dfa-identify_2021}.
The task description, which is also given as the task description
to $L^*LM$ is provided in the appendix (Fig~\ref{fig:task_prompt}). As in our motivating example, two
demonstrations are provided illustrating the task. We instantiate several variants of $L^*LM$ varying the following factors:

\begin{enumerate}[noitemsep,leftmargin=1em]
    \vspace{-1em}
    \item \textbf{all\_rules:} Boolean determining if the third rule of the task prompt is replaced with $\langle$ unknown $\rangle$.
    \item \textbf{allow\_unsure:} Boolean determining if the response grammar includes the ``unsure'' output.
    \item \textbf{use\_L*:} Boolean determining if $L^*$ is used for the DFA-learner or
    \texttt{guess\_dfa\_VL} using SAT.
    \item \textbf{query\_budget}: Number of queries allowed to the LLM per iteration $\in [0, 32]$. 
\end{enumerate}

The results are illustrated in
Figures~\ref{fig:lstar_results}~and~\ref{fig:dfa_identify_results} for \texttt{use\_L*}
= false and true, respectively. These figures plot the minimum energy, $U$, found at
each iteration of DISS. Each figure is broken into four quadrants varying
whether all the rules are provided and whether the unsure response is included
in the grammar. The color of the line indicates the number of queries allowed
per DISS iteration. Finally, the red dotted line corresponds to the energy of the ground truth DFA
shown in Fig~\ref{fig:target-dfa}.

\textbf{RQ1: Natural language prompts improve performance:}
First, observe that with $0$ query budget, we revert to the original DISS experiment as described in~\cite[]{Vazquez-Chanlatte:EECS-2022-107}, i.e., no natural language assistance. 
Second, the DFA conjectured during the first iteration of DISS provides no labeled examples to the DFA learner. Thus, this corresponds to the performance of the learner with no demonstrations.
Studying Figures~\ref{fig:lstar_results}~and~\ref{fig:dfa_identify_results},
we  see that the $0$ query budget runs fail to find DFAs with equal or
lower energy to the ground-truth DFA before the maximum number of DISS
iterations is reached. Conversely, we see that the more both $L^*$ and the
CE backend are allowed to query the LLM, the lower the final
energy tends to be. Further analyzing the learned DFAs, we see that many of the
runs learn the exactly correct DFA, while others learn variants that are nearly
indistinguishable given the demonstration and the particular environment.

\textbf{RQ2: Analyzing the demonstrations is still required to learn the correct DFA:}
Somewhat surprisingly, even in the setting where all rules are provided, we
see that 5 to 15 DISS iterations are still required to learn a good DFA.
As with the Tomita languages, the natural language prompt is not enough and the multi-modal
nature of $L^*LM$ is indeed useful for learning correct DFAs.

\textbf{RQ3: Including more of the task description improves performance:}
Comparing the top and bottom rows of
Figures~\ref{fig:lstar_results}~and~\ref{fig:dfa_identify_results}, we see that
including all rules has small effect on the CE backend
but results in a substantial improvement for the L$^*$ backend. 

\textbf{RQ4: Allowing unsure responses improves performance:}
Comparing the left and right columns of
Figures~\ref{fig:lstar_results}~and~\ref{fig:dfa_identify_results}, we see a clear
improvement in performance when allowing the LLM to respond unsure. Again, the
improvement is particularly noticeable in the L* setting. Analyzing the LLM
responses, this seems to be because L* queries about words to directly 
determine transitions between states as opposed to directly considering
what is relevant between the remaining set of consistent small sized DFAs.
This leads to queries that are largely inconsequential with a higher risk
that a hallucination will lead to a larger than necessary DFA. Further,
this results in DISS providing a correcting labeled example that was
unnecessary given the size prior.

\textbf{RQ5: SAT-based candidate elimination outperforms L$^*$:}
Finally, we observe that the SAT based CE backend
which only queries distinguishing words systematically performs equal to 
or better than the $L^*$ backend -- where the $L^*$ backend often converges just above the red line. This is particularly striking in the treatments
that do not include all the rules or disallow unsure responses.
As with the analysis of unsure responses, this seems to be due to
$L^*$ asking queries with less utility and thus hallucinations
have a larger comparative downside.

Lastly, we consider the effects of changing the output modality of the LLM. We allow the LLMs to output code instead of the context-free-grammar. Samples of the output programs are provided in the appendix. The results are similar to the bottom left graph in Figure~\ref{fig:lstar_results}. The resulting programs are never able to help learn an adequate DFA. Analysis of the chains-of-thought reveal that this is because even less reasoning for the primary task is used in this setting. Instead of focusing on determining whether an example is positive or negative, the LLM quickly makes a superficial decision on the current query and instead focuses its attention on describing all it understands as a program. We also note that in this interactive setting, arbitrary code execution leaves us vulnerable to code injection attacks and mitigation strategies should be employed (e.g. sandboxes).



\bibliography{refs}

\clearpage

\appendix


\section{Demonstration Informed Specification Search (DISS) Overview}
\label{ref:diss_overview}

DISS is a variant of maximum causal entropy inverse reinforcement learning~\cite[]{DBLP:phd/us/Ziebart18},
that requires: 
\begin{enumerate}[leftmargin=1.6em]
\item A concept sampler which returns a concept (here a DFA) consistent with a hypothesized set of labeled examples.
\item A planner which given a concept (here a DFA) returns an entropy regularized policy for it.
\end{enumerate}

DISS proceeds in a loop in which it (i) proposes a candidate DFA from a
set of hypothesized of set of examples, (ii) analyzes the demonstrations using
the entropy regularized policy to hypothesize new labeled examples, and (iii)
adds these new labeled examples into an example buffer which is used to
generate the labeled examples for the next round. The goal of the loop is to minimize
the \textbf{energy} (description complexity) of the DFA plus the demonstration given its corresponding
policy, 
\[
  U(D, \trc) = -\log \Pr(\trc \mid \pi_D) + \lambda\cdot \size(D),
\]
where $\lambda$ is set based on the a-priori expectation of $\size(D)$.

\section{Tomita Grammars}
\label{ref:tomita_exps}

For our first experiment, we use $L^*LM$ as a membership oracle to learn DFAs representing the 7 Tomita Grammars. 
These simple languages are a common benchmark for studying automata learning~\cite[]{pmlr-v80-weiss18a}. 
We ask 30 membership queries for each grammar to GPT3.5-Turbo and GPT4-Turbo. We repeat each experiment for both the original $L^\star$~\cite[]{DBLP:journals/iandc/Angluin87} learning algorithm and $L^*LM$ which allows for extended membership queries. 
The hallucination rates observed $(\frac{\text{\# of incorrect queries}}{\text{total \# of queries}})$ ranged from 3.33\% to 100\% with a median of 33.33\%. 
In all grammars, no model was able to completely avoid hallucination, including a few experiments with GPT-4o. Full experiment details and prompts are provided in the appendix.
Similar hallucination rates were also observed with Mixtral-8x7B-v0.1. These results provide two key takeaways: (i) Allowing LLMs to respond \texttt{unsure} reduces hallucination rates by up to 10\%;, and (ii) despite the reduction, we are unable to eliminate hallucinations and so $L^*LM$ cannot strictly rely on an LLM to serve as an oracle. 
These results demonstrate the need for a grounded supervisory signal and motivate our additional experiment (the running example)  which provides for learning through multiple modalities.

\section{Guessing DFAs as an active version spacing learner}

\begin{center}  
  \begin{minipage}{0.75\linewidth}
    \begin{algorithm}[H]
      \caption{\texttt{guess\_dfa\_VL}
      $(\Sigma, X_+, X_-,$query\_budget$)$\label{alg:version_space_learner}}
      \begin{algorithmic}[1]
      \FOR {$t=1\ldots$ query\_budget}
        \STATE $D_1, D_2 \gets $ \texttt{find\_minimal\_dfas}$(\Sigma, X_+, X_-, 2)$
        \STATE word $\sim L[D_1] \ominus L[D_2]$
        \STATE label $\gets$ M(word)
        \IF{label = true}
          \STATE $X_+ \gets$ word
        \ELSIF{label = false}
          \STATE $X_- \gets$ word
        \ENDIF
      \ENDFOR
      \RETURN \texttt{find\_minimal\_dfas}$(\Sigma, X_+, X_-, 1)$
      \end{algorithmic}
    \end{algorithm}
  \end{minipage}
\end{center}

\section{Psuedo code for L*LM}
\label{ref:appen_psuedo_code}

\begin{python}    
# LLM wrapper prompted with task
oracle = ...
# SAT based DFA identification
find_dfa = ...
# finds word is language sym difference
distinguishing_query = ... 

def guess_dfa(positive, negative)->DFA:
  # 1. Ask membership queries that 
  #  distinguish remaining candidates.
  # Similar process done in 
  #  equivalence query for L* backend
  for _ in range(QUERY_BUDGET):
    word = distinguishing_query(
      positive, negative, alphabet)
    label = oracle(word)
    if label is True:
      positive.append(word)
    elif label is False: 
      negative.append(word)
    else: # idk case
      assert label is None

    # 2. Return minimal consistent DFA.
    return find_dfa(positive, 
      negative, alphabet)

def main():
  diss = DISS()
  positive, negative = [], []
  min_nll, best = float('inf'), None
  while unsatisfied:  # DISS loop
    candidate = guess_dfa(positive, 
      negative, oracle)
    # Compute counterfactual based on
    # on gradient of demonstration nll 
    #  (i.e., surprisal).
    positive, negative, nll = diss.send(
      candidate)
    if nll < min_nll:
      min_nll, best = nll, candidate
  return best
\end{python}

\section{Tomita Grammars}
\subsection{Detailed Results}
The following tables summarize the results of the Tomita Grammar's experiment.

\begin{table}[H]
\centering
\begin{tabular}{|l|c|c|}
\hline
 & Correct & Incorrect \\ \hline
Tomita 1 & 23 & 7 \\ 
Tomita 2 & 29 & 1 \\ 
Tomita 3 & 0 & 30 \\ 
Tomita 4 & 12 & 18 \\ 
Tomita 5 & 20 & 10 \\ 
Tomita 6 & 20 & 10 \\ 
Tomita 7 & 0 & 30 \\ \hline
\end{tabular}
\caption{Active learning of DFAs representing the Seven Tomita Grammars using $L^*LM$ with the original $L^*$ learning algorithm. Results are identical with both \texttt{gpt-3.5-turbo-0125} and \texttt{gpt-4-turbo}.}
\label{tab:incorrect_dfa_no_unsure}
\end{table}

\begin{table}[H]
\centering
\begin{tabular}{|l|c|c|c|}
\hline
 & Unsure & Correct & Incorrect \\ \hline
Tomita 1 & 1 & 23 & 7 \\ 
Tomita 2 & 1 & 29 & 1 \\ 
Tomita 3 & 3 & 0 & 30 \\ 
Tomita 4 & 1 & 12 & 18 \\ 
Tomita 5 & 2 & 20 & 10 \\ 
Tomita 6 & 2 & 20 & 10 \\ 
Tomita 7 & 3 & 0 & 30 \\ \hline
\end{tabular}
\caption{Active learning of DFAs representing the Seven Tomita Grammars using $L^*LM$ and the extended membership variant which includes an unsure option. Results are identical with both \texttt{gpt-3.5-turbo-0125} and \texttt{gpt-4-turbo}.}
\label{tab:incorrect_dfa_with_unsure}
\end{table}

\paragraph{Prompts\\}

The following meta-prompt was used for the treatment which allows an unsure option.

\begin{lstlisting}[language={},frame=single, basicstyle=\ttfamily, framexleftmargin=10pt, xleftmargin=20pt]
The following is a description of a rule for labeling a 
sequence of ones and zeros as good (accepted) or 
bad (rejected).


{rule}


According to the description, respond "true" if the sequence 
is good and "false" if the sequence is bad. If you are unsure 
or do not know the answer, respond "unsure". 
Do not respond with anything else.
\end{lstlisting}

For the treatment which does not allow "unsure", simply omit the second-to-last sentence that describes unsure in the meta-prompt.
\\

For each of the seven Tomita Grammars, we substitute the following prompts into the curly braces of the meta-prompt and query the LLM with the resultant prompt. The prompts we show here were human generated. We also repeated the same experiment with ChatGPT-4o generated prompts from providing the examples and and asking for a natural language explanation of the pattern demonstrated, and finally, we also ask for paraphrased (explained differently) prompts. We found that in any case there was no difference in the downstream $L^*LM$ results.
\\

\noindent
Tomita Grammar One:
\begin{lstlisting}[language={},frame=single, basicstyle=\ttfamily, framexleftmargin=10pt, xleftmargin=20pt]
The sequence should only contain the token '1'.

Seeing any other token should result in rejecting the sequence.

Good examples:
- 1
- 1,1
- 1,1,1
- 1,1,1,1

Bad examples:
- 0
- 1,0
- 0,1
- 1,1,0
\end{lstlisting}

\noindent
Tomita Grammar Two:
\begin{lstlisting}[language={},frame=single, basicstyle=\ttfamily, framexleftmargin=10pt, xleftmargin=20pt]
Only accept sequences that are repetitions of 1,0.

Good examples:
- 1,0
- 1,0,1,0
- 1,0,1,0,1,0
- 1,0,1,0,1,0,1,0

Bad examples
- 1
- 1,0,1
- 0,1,0
- 1,0,1,0,0
\end{lstlisting}

\noindent
Tomita Grammar Three:
\begin{lstlisting}[language={},frame=single, basicstyle=\ttfamily, framexleftmargin=10pt, xleftmargin=20pt]
An odd consecutive sequence of 1 should NEVER be later 
followed by an odd consecutive sequence of zeros.

Good examples:
- 1,0,0
- 0,1,1,0,1,0,0
- 1,1,0,0,0
- 0,0,0,1,1,0,0,0

Bad examples
- 1,0
- 0,1,0
- 1,1,1,0,0,0
- 0,0,0,1,1,1,0,0,0
\end{lstlisting}

\noindent
Tomita Grammar Four:
\begin{lstlisting}[language={},frame=single, basicstyle=\ttfamily, framexleftmargin=10pt, xleftmargin=20pt]
The subsequence 0,0,0 never appears, i.e., no three zeros in a 
row.

Good examples:
- 1
- 1,0,0
- 0,0,1
- 1,1,0,0

Bad Examples:
- 0,0,0
- 1,0,0,0
- 0,0,0,1
- 1,1,0,0,0,1,0
\end{lstlisting}

\noindent
Tomita Grammar Five: 
\begin{lstlisting}[language={},frame=single, basicstyle=\ttfamily, framexleftmargin=10pt, xleftmargin=20pt]
There should be an even number of zeros AND an even number of 
ones.

Good examples:
- 1,1
- 0,0,1,1
- 0,0,1,1,0,0
- 1,1,1,1,0,0

Bad Examples:
- 0,0,0
- 1,0,0,0
- 1,0,0,1
- 0,1,0,1,1
\end{lstlisting}

\noindent
Tomita Grammar Six:
\begin{lstlisting}[language={},frame=single, basicstyle=\ttfamily, framexleftmargin=10pt, xleftmargin=20pt]
The difference between the number of zeros and the number of 
ones is a multiple for 4.

Good examples:
- 1,0
- 0,1
- 0,1,1,1,1
- 0,1,0,1,1,1,1

Bad Examples:
- 1
- 0
- 0,1,1
- 0,1,0,1,1
\end{lstlisting}

\noindent
Tomita Grammar Seven:
\begin{lstlisting}[language={},frame=single, basicstyle=\ttfamily, framexleftmargin=10pt, xleftmargin=20pt]
The sequence 0,1 may appear at most once in the sequence.

Good examples:
- 1
- 0,1
- 0,0,1,0
- 1,0,0

Bad examples:
- 0,1,0,1
- 1,0,1,0,1
- 0,1,1,0,1
- 0,1,0,0,1
\end{lstlisting}

\section{Task Prompt}
\label{fig:task_prompt}
\begin{verbatim}
A robot is operating in a grid world and can visit
four types of tiles: {red, yellow, blue, green}. They
correspond to lava (red), recharging (yellow), water
(blue), and drying (green) tiles. The robot is to
visit tiles according to some set of rules. This will
be recorded as a sequence of colors.

Rules include:

1. The sequence must contain at least one yellow tile,
   i.e., eventually recharge. 
2. The sequence must not contain any red tiles, i.e.,
   lava must be avoided at all costs.
3. If blue is visited, then you must visit green
   *before* yellow, i.e., the robot 
   must dry off before recharging.


A positive example must conform to all rules.
Further note that repeated sequential colors can be replaced with a single
instance.

For example:
1. [yellow,yellow,blue] => [yellow, blue]
2. [red,red,blue,green,green,red] => [red,blue,green,red]
3. [blue,blue,blue] => [blue]
\end{verbatim}

\end{document}